\documentclass[11pt]{article}
\usepackage{acl2016}
\usepackage{times}
\usepackage{url}
\usepackage{latexsym}
\usepackage{booktabs}
\usepackage{multirow}
\usepackage{amsmath}
\usepackage{amssymb}
\usepackage{array}
\usepackage{algorithm}
\usepackage[noend]{algpseudocode}
\usepackage{booktabs}
\usepackage{tikz}
\usepackage{pgfplots}
\usepackage{subcaption}

\aclfinalcopy 

\title{
Log-linear Combinations of Monolingual and Bilingual\\ Neural Machine Translation Models for Automatic Post-Editing
}

\author{Marcin Junczys-Dowmunt \and Roman Grundkiewicz\\
  Adam Mickiewicz University in Pozna\'{n}\\
  ul. Umultowska 87, 61-614 Pozna\'{n}, Poland  \\
  {\tt \{junczys,romang\}@amu.edu.pl} \\}

\date{}

\begin{document}
\maketitle
\begin{abstract}
This paper describes the submission of the AMU (Adam Mickiewicz University) team to the Automatic Post-Editing (APE) task of WMT 2016. We explore the application of neural translation models to the APE problem and achieve good results by treating different models as components in a log-linear model, allowing for multiple inputs (the MT-output and the source) that are decoded to the same target language (post-edited translations).
A simple string-matching penalty integrated within the log-linear model is used to control for higher faithfulness with regard to the raw machine translation output.
To overcome the problem of too little training data, we generate large amounts of artificial data.
Our submission improves over the uncorrected baseline on the unseen test set by -3.2\% TER and +5.5\% BLEU and outperforms any other system submitted to the shared-task by a large margin.
\end{abstract}

\section{Introduction}

This paper describes the submission of the AMU (Adam Mickiewicz University) team to the Automatic Post-Editing (APE) task of WMT 2016. Following the APE shared task from WMT 2015 \cite{bojar2015wmt}, the aim is to test methods for correcting errors produced by an unknown machine translation system in a black-box scenario. The organizers provide training data with human post-edits, evaluation is carried out part-automatically using TER \cite{snover2006study} and
BLEU \cite{Papineni:2002:BMA:1073083.1073135}, and part-manually. 

We explore the application of neural translation models to the APE task and investigate a number of aspects that seem to lead to good results:

\begin{itemize}
 \item Creation of artificial post-edition data that can be used to train the neural models;
 \item Log-linear combination of monolingual and bilingual models in an ensemble-like manner;
 \item Addition of task-specific features in a log-linear model that allow to control for faithfulness of the automatic post-editing output with regard to the input, otherwise a weakness of neural translation models.
\end{itemize}

According to the automatic evaluation metrics used for the task, our system is ranked first among all submission to the shared task.

\section{Related work}

\subsection{Post-Editing}
\label{sec:related:post}
State-of-the-art APE systems follow a monolingual approach firstly proposed by \newcite{simard2007statistical} who trained a phrase-based SMT system on machine translation output and its post-edited versions.
\newcite{bechara2011statistical} proposed a ``source-context aware'' variant of this approach: automatically created word alignments are used to create a new source language which consists of joined MT-output and source token pairs. The inclusion of source-language information in that form is shown to be useful to improve the automatic post-editing results \cite{bechara2012evaluation,chatterjee2015exploring}. The quality of the word alignments plays an important role for this methods, as shown for instance by \newcite{pal2015usaar}.

A number of techniques have been developed to improve PB-SMT-based APE systems, e.g.~approaches relying on phrase-table filtering techniques and specialized features. \newcite{chatterjee2015fbk} propose a pipeline where the best language model and pruned phrase table are selected through task-specific dense features. The goal was to overcome data sparsity issues.

The authors of the Abu-MaTran system (no publication, see \newcite{bojar2015wmt}) incorporate sentence-level classifiers in a post-processing step which choose between the given MT output or an automatic post-edition coming from a PB-SMT APE system.
Their most promising approach consists of a word-level recurrent neural network sequence-to-sequence classifier that marks each word of a sentence as \textit{good} or \textit{bad}.
The output with the lower number of \textit{bad} words is then chosen as the final post-editing answer. We believe this work to be among the first to apply (recurrent) neural networks to the task of automatic post-editing. 

Other popular approaches rely on rule-based components \cite{wisniewski2015why,bechara2012evaluation} which we do not discuss here.

\subsection{Neural machine translation}
We restrict our description to the recently popular encoder-decoder models, based on recurrent neural networks (RNN).  

An LSTM-based encoder-decoder model was introduced by \newcite{sutskever2014sequence}. Here the source sentence is encoded into a single continuous vector, the final state of the source LSTM-RNN. Once the end-of-sentence marker has been encoded, the network generates a translation by sampling the most probable translations from the target LSTM-RNN which keeps its state based on previous words and the source sentence state. 

\newcite{bahdanau2014neural} extended this simple concept with bidirectional source RNNs  \cite{cho2014learning} and the so-called soft-attention model. The novelty of this approach and its improved performance compared to \newcite{sutskever2014sequence} came from the reduced reliance on the source sentence embedding which had to convey all information required for translation in a single state. Instead, attention models learn to look at particular word states at any position within the source sentence. This makes it also easier for these models to learn when to make copies, an important aspect for APE. We refer the reader to \newcite{bahdanau2014neural} for a detailed description of the discussed models.
At the time of writing, no APE systems relying on neural translation models seem to have been published.\footnote{An accepted ACL 2016 paper is scheduled to appear: Santanu Pal, Sudip Kumar Naskar, Mihaela Vela and Josef van Genabith. A Neural Network based Approach to Automated Post-Editing. Proceedings of the 54th Annual Meetings of the Association for Computational Linguistics, August 2016.} 

\section{Data and data preparation}

\subsection{Used corpora}

It was explicitly permitted to use additional data while preparing systems for the APE shared task. We made use of the following resources:

\begin{enumerate}
 \item The official training and development data provided by the APE shared task organizers, consisting of 12,000 training triplets\footnote{A triplet consists of the English source sentence, a German machine translation output, and the German manually post-edited correction of that output.} and 1,000 development set triplets. In this paper we report our results for the 1,000 sentences of development data, and selected results on the unseen test data as provided by the task organizers. 
 \item The domain-specific English-German bilingual training data admissible during the WMT-16 shared task on IT-domain translation;
 \item All other parallel English-German bilingual data admissible during the WMT-16 news translation task;
 \item The German monolingual Common Crawl corpus admissible for the WMT-16 news translation and IT translation tasks.
\end{enumerate}

\subsection{Pre- and post-processing}

The provided triplets have already been tokenized, the tokenization scheme seems to correspond to the Moses \cite{koehn2007moses} tokenizer without escaped special characters, so we re-apply escaping. All other data is tokenized with the Moses tokenizer with standard settings per language. We truecase the data with the Moses truecaser. 

To deal with the limited ability of neural translation models to handle out-of-vocabulary words we split tokens into subword units, following \newcite{sennrich2015neural}. 

Subword units were learned using a 
modified version of the byte pair encoding (BPE) compression algorithm \cite{gage1994new}. 
\newcite{sennrich2015neural} modified the algorithm to work on character level instead of on bytes.
The most frequent pairs of characters are iteratively replaced by a new character 
sequence created by merging the pairs of existent sequences.
Frequent words are thus represented by single symbols and infrequent ones 
are divided into smaller units. The final size of the vocabulary is equal to the sum of merge operations and the number of initial characters. This method effectively reduces the number of unknown words to zero, as characters are always available as the smallest fall-back units. \newcite{sennrich2015neural} showed that this method can deal with German compound nouns (relieving us from applying special methods to handle these) as well as transliterations for Russian-English.

This seems particularly useful in the case of APE, where we do not wish the neural models to ``hallucinate'' output when encountering unknown tokens. A faithful transliteration is more desirable. We chose vocabularies of 40,000 units per language. For German MT output and post-edited sentences we used the same set of subword units. 

\section{Artificial post-editing data}

The provided post-editing data is orders of magnitude too small to train our neural models, and even with the in-domain training data from the IT translation task, we quickly see overfitting effects for a first English-German translation system. Inspired by \newcite{2015arXiv151106709S} --- who use back-translated monolingual data to enrich bilingual training corpora --- we decided to create artificial training triplets. 

\begin{table*}[t]
\begin{center}\renewcommand{\arraystretch}{0.9}
\begin{tabular}{lrcccc}
\toprule
   Data set & Sentences & NumWd & WdSh & NumEr & TER \\
\midrule
  training set        &    12,000 & 17.89 & 0.72 & 4.69 & 26.22 \\
  development set     &     1,000 & 19.76 & 0.71 & 4.90 & 24.81 \\
\midrule
  round-trip.full      & 9,960,000 & 13.50 & 0.58 & 5,72 & 42.02 \\
  round-trip.n10      & 4,335,715 & 15.86 & 0.66 & 5.93 & 36.63 \\
  round-trip.n1       &   531,839 & 20.92 & 0.55 & 5.20 & 25.28 \\
\bottomrule
\end{tabular}
\end{center}
  \caption{Statistics of full and filtered data sets: number of sentences, average number of words, word shifts, errors, and TER score.}  \label{tab:data}
\end{table*}

\subsection{Bootstrapping monolingual data}

We applied cross-entropy filtering \cite{Moore:2010:ISL:1858842.1858883} to the German Common Crawl corpus performing the following steps:

\begin{itemize}
 \item We filtered the corpus for ``well-formed'' lines which start with a capital Unicode letter character and end in an end-of-sentence punctuation mark. We require the line to contain at least 30 Unicode letters. 
 \item The corpus has been preprocessed as described above, including subword units, which may have a positive effect on cross-entropy filtering as they allow to score unknown words.  
 \item Next, we built an in-domain trigram language model \cite{Heafield-estimate} from the German post-editing training data and the German IT-task data, and a similarly sized out-of-domain language model from the Common Crawl data. 
 \item We calculated cross-entropy scores for the first one billion lines of the corpus according to the two language models;
 \item We sorted the corpus by increasing cross-entropy and kept the first 10 million entries for round-trip translation and the top 100 million entries for language modeling.
\end{itemize}

\subsection{Round-trip translation}

For the next step, two phrase-based translation models, English-German and German-English, were created using the admissible parallel training data from the IT task. Word-alignments were computed with fast-align \cite{dyer-chahuneau-smith:2013:NAACL-HLT}, the dynamic-suffix array \cite{fe766d2ec50a42b28a5a10c0f711d14b} holds the translation model. The top 10\% bootstrapped monolingual data was used for language modeling in case of the English-German model, for the German-English translation system the  language model was built only from the target side of the parallel in-domain corpora.\footnote{These models were not meant to be state-of-the-art quality systems. Our main objective was to create them within a few hours.}

The top 1\% of the bootstrapped data have first been translated from German to English and next backwards from English to German. The intermediate English translations were preserved. 
In order to translate these 10 million sentences quickly (twice), we applied small stack-sizes and cube-pruning-pop-limits of around 100, completing the round-trip translation in about 24 hours. 

This procedure left us with 10 million artificial post-editing triplets, where the source German data is treated as post-edited data, the German$\rightarrow$English translated data is the English source, the round-trip translation results are the new uncorrected MT-output. 

\subsection{Filtering for TER}

We hope that a round-trip translation process produces literal translations that may be more-or-less similar to post-edited triplets, where the distance between MT-output and post-edited text is generally smaller than between MT-output and human-produced translations of the same source. Having that much data available, we could continue our filtering process by trying to mimic the TER-statistics of the provided APE training corpus. While TER scores do only take into account the two German language parts of the triplet, it seems reasonable that filtering for better German-German pairs automatically results in a higher quality of the intermediate English part. 

To achieve this, we represented each triplet in the APE training data as a vector of elementary TER statistics computed for the MT-output and the post-edited correction, such as the sentence length, the frequency of edit operations, and the sentence-level TER score. We do the same for the to-be-filtered artificial triplet corpus. The similarity measure is the inverse Euclidean distance over these vector representations. 

In a first step, outliers which diverge from any maximum or minimum value of the reference vectors by more than 10\% were removed. For example, we filtered triplets with post-edited sentences that were 10\% longer than the longest post-edited sentence in the reference.

In the second step, for each triplet from the reference set we select $n$ nearest neighbors. Candidates that have been chosen for one reference set triplet were excluded for the following triplets. If more than the 100 triplets had to be traversed to satisfy the exclusion criterion, less than $n$ or even 0 candidates were selected. Two subsets have been created, one for $n=1$ and one for $n=10$. Table~\ref{tab:data} sets the characteristics of the obtained corpora in relation to the provided training and development data. The smaller set (round-trip.n1) follows the TER statistics of the provided training and development data quite closely, but consists only of 5\% of the artificial triplets. The larger set (round-trip.n10) consists of roughly 43\% of the data, but has weaker TER scores. 

\begin{figure}[t]
\centering
\begin{tikzpicture}
\begin{axis}[
y tick label style={/pgf/number format/.cd, fixed, fixed zerofill, precision=1, /tikz/.cd},
ymajorgrids,
mark options=solid,
width=0.49\textwidth,
legend pos= north east,
thick, table/x index={0},legend pos=south east,
xlabel={$n \times 10000$ iterations}]
\addplot+[mark=none] table[y index = {1}]{progress.dat};
\addplot+[mark=none] table[y index = {2}]{progress.dat};
\draw[ultra thin, dashed] (axis cs:30,\pgfkeysvalueof{/pgfplots/ymin}) -- (axis cs:30,\pgfkeysvalueof{/pgfplots/ymax});
 \legend{mt-pe, src-pe, baseline}
\end{axis}
\end{tikzpicture}
\caption{Training progress for mt-pe and src-pe models according to development set; dashed vertical line marks change from training set round-trip.n10 to fine-tuning with round-trip.n1.}\label{fig:progress}
\end{figure}
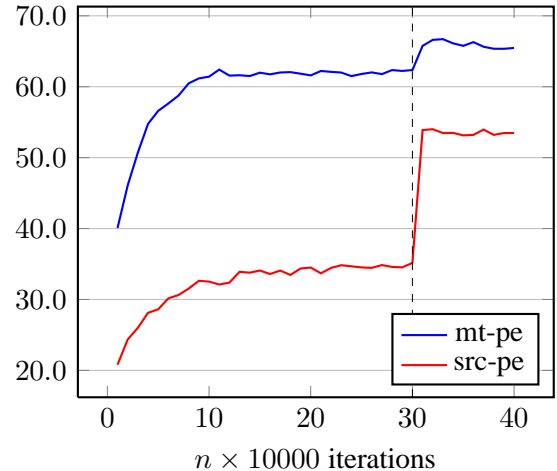

\section{Experiments}

Following the post-editing-by-machine-trans\-la\-tion paradigm, we explore the application of soft-attention neural translation models to post-editing. Analogous to the two dominating approaches described in Section \ref{sec:related:post}, we investigate methods that are purely monolingual as well as a simple method to include source language information in a more natural way than it has been done for phrase-based machine translation. 

The neural machine translation systems explored in this work are attentional encoder-decoder models \cite{bahdanau2014neural}, which have been trained with Nematus\footnote{\url{https://github.com/rsennrich/nematus}}.
We used mini-batches of size 80, a maximum sentence length of 50, word embeddings of size 500, and hidden layers of size 1024. Models were trained with Adadelta \cite{DBLP:journals/corr/abs-1212-5701}, reshuffling the corpus between epochs. As mentioned before tokens were split into subword units, 40,000 per language. 
For decoding, we used AmuNMT\footnote{\url{https://github.com/emjotde/amunmt}}, our C++/CUDA decoder for NMT models trained with Nematus with a beam size of 12 and length normalization. 

\subsection{MT-output to post-editing}

We started training the monolingual MT-PE model with the MT and PE data from the larger artificial triplet corpus (round-trip.n10). The model has been trained for 4 days, saving a model every $10,000$ mini-batches. Quick convergence can be observed for the monolingual task and we switched to fine-tuning after the 300,000-th iteration with a mix of the provided training data and the smaller round-trip.n1 corpus. The original post-editing data was oversampled 20 times and concatenated with round-trip.n1. 

This resulted in the performance jump shown in Figure~\ref{fig:progress} (mt$\rightarrow$pe, blue). 
Training were continued for another 100,000 iterations and stopped when overfitting effects became apparent. Training directly with the smaller training data without the initial training on round-trip.n10 lead to even earlier overfitting. 

Entry mt$\rightarrow$pe in Table~\ref{tab:dev} contains the results of the single-best model on the development set which outperforms the baseline significantly. Models for ensembling are selected among the periodically saved parameter dumps of one training run. An ensemble mt$\rightarrow$pe$\times 4$ consisting of the four best models shows only modest improvements over the single model. The same development set has been used to select the best-performing models, results may therefore be slightly skewed.

\begin{table}[t]
\centering
\setlength{\tabcolsep}{5pt}
 \begin{tabular}{lcc} \toprule
  System & TER & BLEU \\ \midrule
  Baseline (mt) & 25.14 & 62.92 \\ \midrule
  mt$\rightarrow$pe & 23.37 & 66.71 \\
  mt$\rightarrow$pe$\times 4$ & 23.23 & 66.88 \\  \midrule
  src$\rightarrow$pe & 32.31 & 53.89 \\
  src$\rightarrow$pe$\times 4$ & 31.42 & 55.41 \\  \midrule
  mt$\rightarrow$pe$\times 4$ / src$\rightarrow$pe$\times 4$$^*$ & 22.38 & 68.07 \\
  mt$\rightarrow$pe$\times 4$ / src$\rightarrow$pe$\times 4$ / pep$^*$ & \textbf{21.46} & \textbf{68.94} \\ \bottomrule
  \end{tabular}
  \caption{Results on provided development set. Best-performing models have been chosen based on this development set. Systems marked with $^*$ have weights tuned on the same development set.}\label{tab:dev}
\end{table}

\subsection{Source to post-editing}

We proceed similarly for the English-German NMT training. When fine-tuning with the smaller corpus with oversampled post-editing data, we also add all in-domain parallel training data from the IT-task, roughly 200,000 sentences. Fine-tuning results in a much larger jump than in the monolingual case, but the overall performance of the NMT system is still weaker than the uncorrected MT-baseline.  

As for the monolingual case, we evaluate the single-best model (src$\rightarrow$pe) and an ensemble (src$\rightarrow$pe$\times 4$) of the four best models of a training run. The src$\rightarrow$pe$\times 4$ system is not able to beat the MT baseline, but the ensemble is significantly better than the single model. 

\subsection{Log-linear combinations and tuning}

AmuNMT can be configured to accept different inputs to different members of a model ensemble as long as the target language vocabulary is the same. We can therefore build a decoder that takes both, German MT output and the English source sentence, as parallel input, and produces post-edited German as output. Since once the input sentence has been provided to a NMT model it essentially turns into a language model, this can be achieved without much effort. In theory an unlimited number of inputs can be combined in this way without the need of specialized multi-input training procedures \cite{DBLP:journals/corr/ZophK16}.\footnote{Which are still worth investigating for APE and likely to yield better results.} 

In NMT ensembles, homogeneous models are typically weighted equally. Here we combine different models and equal weighting does not work. Instead, we treat each ensemble component as a feature in a traditional log-linear model and perform weighting as parameter tuning with Batch-Mira \cite{Cherry:2012:BTS:2382029.2382089}. AmuNMT can produce Moses-compatible n-best lists and we devised an iterative optimization process similar to the one available in Moses. We tune the weights on the development set towards lower TER scores; two iterations seem to be enough. When ensembling one mt$\rightarrow$pe model and one src$\rightarrow$pe model, the assigned weights correspond roughly to 0.8 and 0.2 respectively.
The linear combination of all eight models (mt$\rightarrow$pe$\times 4$ / src$\rightarrow$pe$\times 4$) improves quality by 0.9 TER and 1.2 BLEU, however, weights were tuned on the same data.

\subsection{Enforcing faithfulness}

We extend AmuNMT with a simple Post-Editing Penalty (PEP).
To ensure that the system is fairly conservative --- i.e. the correction process does not introduce too much new material --- every word in the system's output that was
not seen in its input incurs a penalty of -1.  

During decoding this is implemented efficiently as a matrix of dimensions batch size $\times$ target vocabulary size where all columns that match source words are assigned $0$ values, all other words $-1$. This feature can then be used as if it was another ensemble model and tuned with the same procedure as described above.  

PEP introduces a precision-like bias into the decoding process and is a simple means to enforce a certain faithfulness with regard to the input via string matching. This is not easily accomplished within the encoder-decoder framework which abstracts away from any string representations. A recall-like variant (penalize for missing input words in the output) cannot be realized at decode-time as it is not known which words have been omitted until the very end of the decoding process. This could only work as a final re-ranking criterion, which we did not explore in this paper. The bag-of-words approach grants the NMT model the greatest freedom with regard to reordering and fluency for which these models seem to be naturally well-suited. 

As before, we tune the combination on the development set. The resulting system (mt$\rightarrow$pe$\times 4$ / src$\rightarrow$pe$\times 4$ / pep) can again improve post-editing quality. We see a total improvement of -3.7\% TER and +6.0\% BLEU over the given MT baseline on the development set. The log-linear combination of different features improves over the purely monolingual ensemble by -1.8\% TER and +2.1\% BLEU.

\section{Final results and conclusions}

\begin{table}[t]
\centering
\centering 
 \begin{tabular}{lcc} \toprule
  System & TER & BLEU \\ \midrule
mt$\rightarrow$pe$\times 4$ / src$\rightarrow$pe$\times 4$ / pep & \textbf{21.52} & \textbf{67.65}\\
mt$\rightarrow$pe$\times 4$ (contrastive) & \bf 23.06 & \bf 66.09 \\ 
FBK	& 23.92 & 64.75 \\
USAAR	& 24.14 & 64.10\\
CUNI & 24.31 & 63.32\\
Standard Moses (baseline 2) & 24.64 & 63.47 \\
Uncorrected MT (baseline 1) & 24.76 & 62.11 \\ 
DCU & 26.79 & 58.60\\
JUSAAR & 26.92 & 59.44\\
  \bottomrule
  \end{tabular}
  \caption{Results on unseen test set in comparison to other shared task submissions as reported by the task organizers. For submissions by other teams we include only their best result.}\label{tab:test}
\end{table}

We submitted the output of the last system (mt$\rightarrow$pe$\times 4$ / src$\rightarrow$pe$\times 4$ / pep) as our final proposition for the APE shared task, and mt$\rightarrow$pe$\times 4$ as a contrastive system. Table~\ref{tab:test} contains the results on the unseen test set for our two systems (in bold) and the best system of any other submitting team as reported by the task organizers (for more details and manually judged results --- which were not yet available at the time of writing --- see the shared task overview paper). Results are sorted by TER from best to worse.
For our best system, we see improvements of -3.2\% TER and +5.5\% BLEU over the unprocessed baseline~1 (uncorrected MT), and -1.5\% TER and +1.5\% BLEU over our contrastive system. The organizers also provide results for a standard phrase-based Moses set-up (baseline 2) that can hardly beat baseline~1 (-0.1\% TER, +1.4\% BLEU). Both our systems outperform the next-best submission by large margins. In the light of these last results, our system seems to be quite successful. 

We could demonstrate the following:
\begin{itemize}
 \item Neural machine translation models can be successfully applied to APE;
 \item Artificial APE triplets help against early overfitting and make it possible to overcome the problem of too little training data;
 \item Log-linear combinations of neural machine translation models with different input languages can be used as a method of combining MT-output and source data for APE to positive effects;
 \item Task specific features can be easily integrated into the log-linear models and can control the faithfulness of the APE results. 
\end{itemize}

Future work should include the investigation of integrated multi-source approaches like \cite{DBLP:journals/corr/ZophK16} and better schemes of dealing with overfitting. We also plan to apply our methods to the data of last year's APE task. 
\section{Acknowledgements}

This work is partially funded by the National Science Centre, Poland (Grant No. 2014/15/N/ST6/02330). 

\bibliography{wmt16ape}

\begin{thebibliography}{}

\bibitem[\protect\citename{Bahdanau \bgroup et al.\egroup
  }2015]{bahdanau2014neural}
Dzmitry Bahdanau, Kyunghyun Cho, and Yoshua Bengio.
\newblock 2015.
\newblock Neural machine translation by jointly learning to align and
  translate.
\newblock In {\em Proceedings of the International Conference on Learning
  Representations}, San Diego, CA.

\bibitem[\protect\citename{B{\'e}chara \bgroup et al.\egroup
  }2011]{bechara2011statistical}
Hanna B{\'e}chara, Yanjun Ma, and Josef van Genabith.
\newblock 2011.
\newblock Statistical post-editing for a statistical {MT} system.
\newblock In {\em Proceedings of the 13th Machine Translation Summit}, pages
  308--315, Xiamen, China.

\bibitem[\protect\citename{B{\'e}chara \bgroup et al.\egroup
  }2012]{bechara2012evaluation}
Hanna B{\'e}chara, Rapha{\"e}l Rubino, Yifan He, Yanjun Ma, and Josef van
  Genabith.
\newblock 2012.
\newblock An evaluation of statistical post-editing systems applied to {RBMT}
  and {SMT} systems.
\newblock In {\em Proceedings of COLING 2012}, pages 215--230, Mumbai, India.

\bibitem[\protect\citename{Bojar \bgroup et al.\egroup }2015]{bojar2015wmt}
Ond\v{r}ej Bojar, Rajen Chatterjee, Christian Federmann, Barry Haddow, Matthias
  Huck, Chris Hokamp, Philipp Koehn, Varvara Logacheva, Christof Monz, Matteo
  Negri, Matt Post, Carolina Scarton, Lucia Specia, and Marco Turchi.
\newblock 2015.
\newblock Findings of the 2015 {Workshop on Statistical Machine Translation}.
\newblock In {\em Proceedings of the Tenth Workshop on Statistical Machine
  Translation}, pages 1--46, Lisbon, Portugal. Association for Computational
  Linguistics.

\bibitem[\protect\citename{Chatterjee \bgroup et al.\egroup
  }2015a]{chatterjee2015fbk}
Rajen Chatterjee, Marco Turchi, and Matteo Negri.
\newblock 2015a.
\newblock The {FBK} participation in the {WMT15} automatic post-editing shared
  task.
\newblock In {\em Proceedings of the Tenth Workshop on Statistical Machine
  Translation}, pages 210--215, Lisbon, Portugal. Association for Computational
  Linguistics.

\bibitem[\protect\citename{Chatterjee \bgroup et al.\egroup
  }2015b]{chatterjee2015exploring}
Rajen Chatterjee, Marion Weller, Matteo Negri, and Marco Turchi.
\newblock 2015b.
\newblock Exploring the planet of the {APEs}: a comparative study of
  state-of-the-art methods for {MT} automatic post-editing.
\newblock In {\em Proceedings of the 53rd Annual Meeting of the Association for
  Computational Linguistics and the 7th International Joint Conference on
  Natural Language Processing}, pages 156--161, Beijing, China. Association for
  Computational Linguistics.

\bibitem[\protect\citename{Cherry and
  Foster}2012]{Cherry:2012:BTS:2382029.2382089}
Colin Cherry and George Foster.
\newblock 2012.
\newblock Batch tuning strategies for statistical machine translation.
\newblock In {\em Proceedings of the 2012 Conference of the North American
  Chapter of the Association for Computational Linguistics: Human Language
  Technologies}, pages 428--436, Stroudsburg, PA, USA. Association for
  Computational Linguistics.

\bibitem[\protect\citename{Cho \bgroup et al.\egroup }2014]{cho2014learning}
Kyunghyun Cho, Bart van Merrienboer, Caglar Gulcehre, Dzmitry Bahdanau, Fethi
  Bougares, Holger Schwenk, and Yoshua Bengio.
\newblock 2014.
\newblock Learning phrase representations using {RNN} encoder--decoder for
  statistical machine translation.
\newblock In {\em Proceedings of the 2014 Conference on Empirical Methods in
  Natural Language Processing}, pages 1724--1734, Doha, Qatar. Association for
  Computational Linguistics.

\bibitem[\protect\citename{Dyer \bgroup et al.\egroup
  }2013]{dyer-chahuneau-smith:2013:NAACL-HLT}
Chris Dyer, Victor Chahuneau, and Noah~A. Smith.
\newblock 2013.
\newblock A simple, fast, and effective reparameterization of {IBM} model 2.
\newblock In {\em Proceedings of the 2013 Conference of the North American
  Chapter of the Association for Computational Linguistics: Human Language
  Technologies}, pages 644--648, Atlanta, Georgia. Association for
  Computational Linguistics.

\bibitem[\protect\citename{Gage}1994]{gage1994new}
Philip Gage.
\newblock 1994.
\newblock A new algorithm for data compression.
\newblock {\em The C Users Journal}, (2):23--38.

\bibitem[\protect\citename{Germann}2015]{fe766d2ec50a42b28a5a10c0f711d14b}
Ulrich Germann.
\newblock 2015.
\newblock Sampling phrase tables for the {Moses} statistical machine
  translation system.
\newblock {\em Prague Bulletin of Mathematical Linguistics}, (1):39--50.

\bibitem[\protect\citename{Heafield \bgroup et al.\egroup
  }2013]{Heafield-estimate}
Kenneth Heafield, Ivan Pouzyrevsky, Jonathan~H. Clark, and Philipp Koehn.
\newblock 2013.
\newblock Scalable modified {Kneser-Ney} language model estimation.
\newblock In {\em Proceedings of the 51st Annual Meeting of the Association for
  Computational Linguistics}, pages 690--696, Sofia, Bulgaria.

\bibitem[\protect\citename{Koehn \bgroup et al.\egroup }2007]{koehn2007moses}
Philipp Koehn, Hieu Hoang, Alexandra Birch, Chris Callison-Burch, Marcello
  Federico, Nicola Bertoldi, Brooke Cowan, Wade Shen, Christine Moran, Richard
  Zens, et~al.
\newblock 2007.
\newblock Moses: Open source toolkit for statistical machine translation.
\newblock In {\em Proceedings of the 45th Annual Meeting of the Association for
  Computational Linguistics}, pages 177--180. Association for Computational
  Linguistics.

\bibitem[\protect\citename{Moore and
  Lewis}2010]{Moore:2010:ISL:1858842.1858883}
Robert~C. Moore and William Lewis.
\newblock 2010.
\newblock Intelligent selection of language model training data.
\newblock In {\em Proceedings of the 48th Annual Meeting of the Association for
  Computational Linguistics}, pages 220--224, Stroudsburg, PA, USA. Association
  for Computational Linguistics.

\bibitem[\protect\citename{Pal \bgroup et al.\egroup }2015]{pal2015usaar}
Santanu Pal, Mihaela Vela, Sudip~Kumar Naskar, and Josef van Genabith.
\newblock 2015.
\newblock {USAAR-SAPE}: An {English}--{Spanish} statistical automatic
  post-editing system.
\newblock In {\em Proceedings of the Tenth Workshop on Statistical Machine
  Translation}, pages 216--221, Lisbon, Portugal. Association for Computational
  Linguistics.

\bibitem[\protect\citename{Papineni \bgroup et al.\egroup
  }2002]{Papineni:2002:BMA:1073083.1073135}
Kishore Papineni, Salim Roukos, Todd Ward, and Wei-Jing Zhu.
\newblock 2002.
\newblock {BLEU}: A method for automatic evaluation of machine translation.
\newblock In {\em Proceedings of the 40th Annual Meeting on Association for
  Computational Linguistics}, ACL '02, pages 311--318, Stroudsburg, PA, USA.
  Association for Computational Linguistics.

\bibitem[\protect\citename{Sennrich \bgroup et al.\egroup
  }2015a]{2015arXiv151106709S}
Rico Sennrich, Barry Haddow, and Alexandra Birch.
\newblock 2015a.
\newblock Improving neural machine translation models with monolingual data.
\newblock {\em arXiv preprint arXiv:1511.06709}.

\bibitem[\protect\citename{Sennrich \bgroup et al.\egroup
  }2015b]{sennrich2015neural}
Rico Sennrich, Barry Haddow, and Alexandra Birch.
\newblock 2015b.
\newblock Neural machine translation of rare words with subword units.
\newblock {\em arXiv preprint arXiv:1508.07909}.

\bibitem[\protect\citename{Simard \bgroup et al.\egroup
  }2007]{simard2007statistical}
Michel Simard, Cyril Goutte, and Pierre Isabelle.
\newblock 2007.
\newblock Statistical phrase-based post-editing.
\newblock In {\em Proceedings of the Conference of the North American Chapter
  of the Association for Computational Linguistics}, pages 508--515, Rochester,
  New York. Association for Computational Linguistics.

\bibitem[\protect\citename{Snover \bgroup et al.\egroup }2006]{snover2006study}
Matthew Snover, Bonnie Dorr, Richard Schwartz, Linnea Micciulla, and John
  Makhoul.
\newblock 2006.
\newblock A study of translation edit rate with targeted human annotation.
\newblock In {\em Proceedings of Association for Machine Translation in the
  Americas}, pages 223--231, Cambridge, Massachusetts.

\bibitem[\protect\citename{Sutskever \bgroup et al.\egroup
  }2014]{sutskever2014sequence}
Ilya Sutskever, Oriol Vinyals, and Quoc~V Le.
\newblock 2014.
\newblock Sequence to sequence learning with neural networks.
\newblock In {\em Advances in Neural Information Processing Systems 27: 28th
  Annual Conference on Neural Information Processing Systems 2014}, pages
  3104--3112, Montreal, Canada.

\bibitem[\protect\citename{Wisniewski \bgroup et al.\egroup
  }2015]{wisniewski2015why}
Guillaume Wisniewski, Nicolas P\'{e}cheux, and Fran\c{c}ois Yvon.
\newblock 2015.
\newblock Why predicting post-edition is so hard? failure analysis of {LIMSI}
  submission to the {APE} shared task.
\newblock In {\em Proceedings of the Tenth Workshop on Statistical Machine
  Translation}, pages 222--227, Lisbon, Portugal. Association for Computational
  Linguistics.

\bibitem[\protect\citename{Zeiler}2012]{DBLP:journals/corr/abs-1212-5701}
Matthew~D. Zeiler.
\newblock 2012.
\newblock {ADADELTA:} an adaptive learning rate method.
\newblock {\em arXiv preprint arXiv:1212.5701}.

\bibitem[\protect\citename{Zoph and Knight}2016]{DBLP:journals/corr/ZophK16}
Barret Zoph and Kevin Knight.
\newblock 2016.
\newblock Multi-source neural translation.
\newblock {\em arXiv preprint arXiv:1601.00710}.

\end{thebibliography}
\bibliographystyle{acl2016}

\end{document}